\documentclass[conference]{IEEEtran}
\IEEEoverridecommandlockouts
\usepackage{cite}
\usepackage{amsmath,amssymb,amsfonts}
\usepackage{algorithmic}
\usepackage{graphicx}
\usepackage{textcomp}
\usepackage{booktabs}
\usepackage{xcolor}
\usepackage{multirow}
\usepackage{url}
\usepackage[table,xcdraw]{xcolor}
\def\BibTeX{{\rm B\kern-.05em{\sc i\kern-.025em b}\kern-.08em
    T\kern-.1667em\lower.7ex\hbox{E}\kern-.125emX}}
\begin{document}

\title{RareDxR1: Autonomous Medical Reasoning for Rare Disease Diagnosis Beyond Human Annotation}

% \and
% \and
% \and
% \and
% \and

\author{
\IEEEauthorblockN{
Deyang Jiang$^{1,2,*}$,
Haoran Wu$^{1,*}$,
Ziyi Wang$^{1}$,
Yiming Rong$^{1}$,
Yunlong Zhao$^{1}$,
Ye Jin$^{3}$, and
Bo Xu$^{1,2}$
\thanks{$^{*}$Equal contribution. Corresponding author: Bo Xu (xubo@ia.ac.cn).}
\thanks{The project is supported by the National High Level Hospital Clinical Research Funding (2025-PUMCH-C-006), and the National Natural Science Foundation of China (No. 62506358).}
\thanks{Accepted to the IEEE International Conference on Multimedia and Expo (ICME) 2026. \copyright~2026 IEEE. Personal use of this material is permitted. Permission from IEEE must be obtained for all other uses, in any current or future media, including reprinting/republishing this material for advertising or promotional purposes, creating new collective works, for resale or redistribution to servers or lists, or reuse of any copyrighted component of this work in other works.}
}

\vspace{0.2cm}

\IEEEauthorblockA{$^{1}$The Key Laboratory of Cognition and Decision Intelligence for Complex Systems, \\Institute of Automation, Chinese Academy of Sciences, Beijing, China}
\IEEEauthorblockA{$^{2}$School of Future Technology, University of Chinese Academy of Sciences, Beijing, China}
\IEEEauthorblockA{$^{3}$Peking Union Medical College Hospital, Beijing, China}

\vspace{0.1cm}

% \IEEEauthorblockA{
% Email: jiangdeyang2022@ia.ac.cn, wuhaoran2018@ia.ac.cn, wangziyi2022@ia.ac.cn, \\
% rongyiming2022@ia.ac.cn, zhaoyunlong2020@ia.ac.cn, jinye10564@pumch.cn, xubo@ia.ac.cn
% }

% \thanks{Corresponding author:Bo Xu (xubo@ia.ac.cn). The project is supported by National High Level Hospital Clinical Research Funding (2025-PUMCH-C-006), and also supported by National Natural Science Foundation of China No. 62506358.}

\thanks{The project is supported by the National High Level Hospital Clinical Research Funding (2025-PUMCH-C-006), and the National Natural Science Foundation of China (No. 62506358).}

}

\maketitle

\begin{abstract}
Rare disease differential diagnosis is a critical yet arduous clinical task, requiring physicians to identify precise phenotypes from complex, unstructured patient symptoms and execute intricate reasoning within a vast search space. However, existing AI approaches typically rely on pipeline-based phenotype extraction or retrieval-augmented generation, which suffer from critical information loss due to predefined ontologies, retrieval bottlenecks, and a lack of diagnostic logic. 
To address these challenges, we introduce RareDxR1, an end-to-end reasoning-centric large language model designed for open-domain rare disease diagnosis directly from unstructured clinical notes.
We design a progressive end-to-end training framework by synergizing knowledge internalization with autonomous evolutionary learning, thereby bypassing reliance on structured phenotypes and closed-set decision-making. To overcome the limitations of RAG and phenotype restriction, we enabled the deep internalization of fragmented rare disease knowledge directly into the model's parameters. Moreover, to bridge the gap between model generation and expert reasoning, we propose Reflection-Enhanced Reasoning Sampling (RERS), a strategy that synthesizes expert-level diagnostic trajectories by learning from failures without human annotation. Additionally, we propose a dual-level curriculum reinforcement learning approach for gradually mastering rare disease diagnosis. Experimental results demonstrate that RareDxR1 achieves a state-of-the-art accuracy across different benchmarks, marking a significant breakthrough in open-domain rare disease diagnosis. Our code and dataset will be publicly available.\footnote{\url{https://github.com/jdy18/RareDxR1}}
\end{abstract}


\section{Introduction}
Rare diseases, affecting less than 1/2000 of the population, encompass 7,000–10,000 types globally and impact approximately 300 million people~\cite{haendel2020many, yicheng2022retrospective}. \textbf{Rare Disease Diagnosis (RDD)—the process of identifying specific diseases from undifferentiated clinical manifestations}—is notoriously difficult. In clinical practice, this task requires a dual-process capability: first, physicians must meticulously parse unstructured patient narratives to identify key phenotypes; second, they must perform complex differential diagnosis, weighing conflicting evidence against multiple conditions. The scarcity of specialized expertise often leads to a 4.8-year average diagnostic delay and frequent misdiagnoses~\cite{evans2018dare, martinez2022epigenomic}.

\begin{figure}[tb] 
    \centering %
    \includegraphics[width=0.4\textwidth]{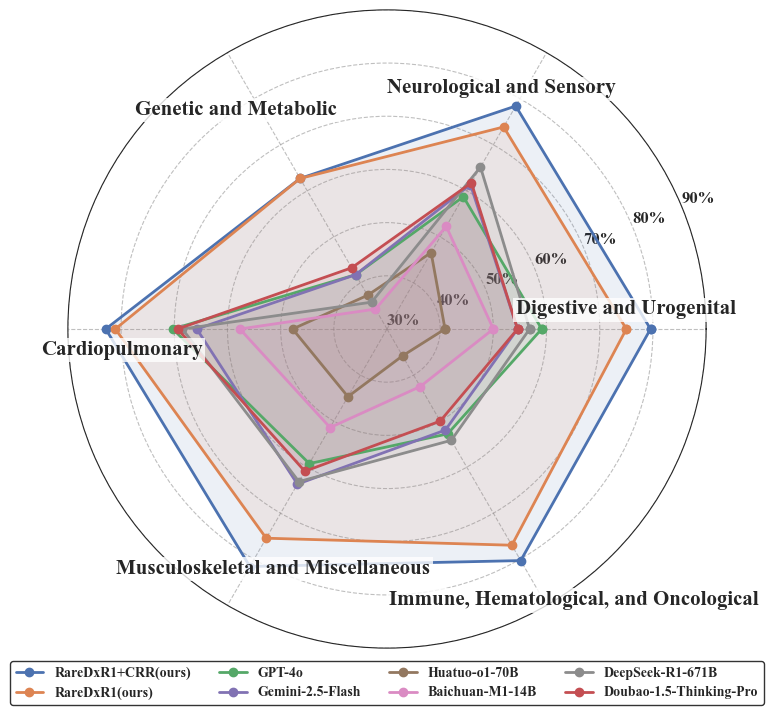} 
    \caption{Top-10 Recall across all rare disease categories in RareArena-Test.} 
    \label{fig:radar} 
\end{figure}

To assist this decision-making, traditional computational methods have long relied on standardized phenotype ontologies~\cite{mao2025phenotype,zhai2023phen2disease,peng2016measuring,pinol2017rarediscovery} or genotypic data~\cite{zhao2020phen2gene, deisseroth2019clinphen, smedley2015phenotypegene} as core features. However, such symbolic inputs depend heavily on manual feature engineering and often overlook critical nuances embedded in unstructured notes. Furthermore, these methods are typically confined to predefined, closed label spaces, making it difficult to recognize unseen diseases or discover potential rare conditions in open-set clinical practice. 

The development of Large Language Models (LLMs) has made open-domain diagnosis directly from raw text possible. However, they currently face two critical gaps: insufficient rare disease knowledge and inadequate clinical reasoning. On the knowledge front, current Medical-LLMs~\cite{wang2025baichuan, wang2023huatuo} lack coverage of long-tail diseases, leading to hallucinations and a failure to utilize structured phenotypic databases~\cite{talapova2023human, amberger2015omim}. The reasoning gap is even more severe. General reasoning models\cite{guo2025deepseek,deepmind2025gemini} excel in formal logic but struggle with the probabilistic nature of clinical diagnosis, while Medical-LLMs lack the expert-level data needed to navigate complex diagnostic paths. Many researchers employ Rejection Sampling (RS)~\cite{ahn2024large} to synthesize Chain-of-Thought data, but it is inefficient here: general models rarely generate correct initial solutions for rare diseases, and discarding incorrect samples wastes valuable opportunities to learn from failures.

To bridge these gaps, we introduce RareDxR1, an end-to-end reasoning-centric large language model designed for open-domain rare disease diagnosis. Unlike previous methods, it performs direct diagnostic inference from unstructured clinical notes, without the rigid phenotype extraction step. Moreover, it achieves expert-level multi-step reasoning capabilities without requiring any human annotation.
Our approach addresses the challenges through four key technical innovations:
\begin{enumerate}
\item To resolve retrieval bottlenecks, we propose a strategy for Knowledge Internalization. This allows the model to encode rare disease facts directly into its parameters, enabling effective "knowledge emergence" without the reliance on external databases typical of phenotype-based or RAG systems.
\item To mitigate the lack of reasoning capacity in data-sparse scenarios, we introduce Reflection-Enhanced Reasoning Sampling. It forces the model to reflect on its own diagnostic failures and correct them. 
\item To overcome the limitations of structured phenotype dependence and closed-set decision-making, we propose an integrated end-to-end training pipeline and derive RareDxR1, which enables direct, open-domain diagnosis from raw clinical notes, achieving SOTA performance on multiple benchmarks.
\item To integrate the advantages of other models and reduce model-specific diagnostic biases, we propose Collaborative Reasoning Refinement (CRR). This is a simple yet effective inference strategy that further improves diagnostic accuracy by integrating insights from external models to refine the initial diagnosis.

\end{enumerate}
% \begin{figure}[t] 
%     \centering %
%     \includegraphics[width=0.5\textwidth]{figs/cot_campare.png} 
%     \caption{Comparison of diagnostic reasoning between RareDxR1 and a conventional medical reasoning model (Huatuo-o1-70B). To intuitively illustrate their differing thought processes on the same complex case, we have summarized the models' free-text reasoning into structured mind trees. RareDxR1 demonstrates a systematic differential diagnosis process, correctly weighing conflicting evidence to arrive at the final diagnosis. In contrast, the conventional model follows a linear reasoning path, fixating on an initial hypothesis and failing to self-correct despite contradictory evidence.} 
%     \label{fig:overall} 
% \end{figure}

\section{Related Works}

\subsection{Rare Disease Diagnosis}
Traditional approaches, whether phenotype-based~\cite{mao2025phenotype,zhai2023phen2disease,peng2016measuring,pinol2017rarediscovery} or gene-based~\cite{zhao2020phen2gene, deisseroth2019clinphen, smedley2015phenotypegene}, heavily rely on manually curated annotations in knowledge bases. Recently, Large Language Models (LLMs) have been applied via retrieval-augmented reasoning~\cite{wu2025integratingchainofthoughtretrievalaugmented}, multi-agent collaboration~\cite{chen2025rareagentsadvancingraredisease}, and tool invocation~\cite{zhao2025deeprare}. While continued medical pre-training~\cite{wang2025baichuan, wang2023huatuo} has yielded gains, these general-purpose models still struggle with the sparse, specialized knowledge and intricate reasoning required for rare disease diagnosis.
\subsection{Reasoning Language Models}
Reinforcement learning (RL) has proven effective for enhancing reasoning, supported by algorithm optimizations like GRPO~\cite{shao2024deepseekmathpushinglimitsmathematical} and DAPO~\cite{yu2025dapoopensourcellmreinforcement}. In the medical domain, recent works such as Med-R1~\cite{lai2025medr1reinforcementlearninggeneralizable}, Med-RLVR~\cite{zhang2025medrlvremergingmedicalreasoning}, and Baichuan-M1~\cite{wang2025baichuan} have utilized RL for alignment. However, these efforts predominantly focus on general medical QA, leaving the significantly more complex challenge of open-domain rare disease differential diagnosis largely unexplored.

\begin{figure*}[tb]
    \centering
    % Placeholder for the figure, as it was not provided in the text
    \includegraphics[width=0.9\textwidth]{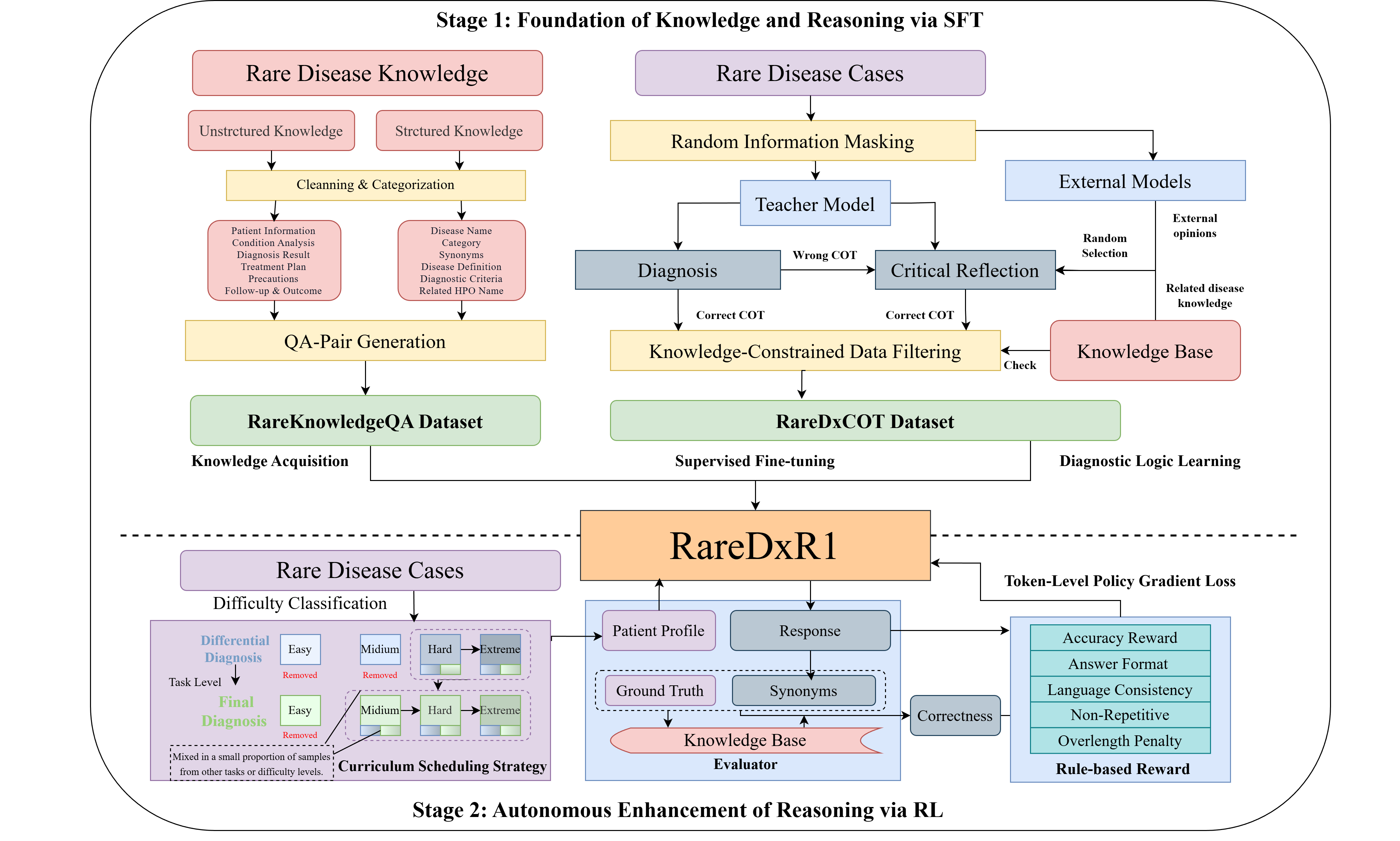}
    \vspace{2pt}  
    \caption{The two-stage training framework for constructing the RareDxR1 model.}
    \label{fig:training_framework}
\end{figure*}

\section{Methods}
To address the challenges of sparse knowledge and complex reasoning, we develop \textbf{RareDxR1} with a multi-stage training framework followed by a collaborative inference strategy.  

In this section, we first present our data engine designed for the Supervised Fine-Tuning stage: \textbf{RareKnowledgeQA} for knowledge integration and \textbf{RareDxCOT} for expert-level reasoning synthesis via RERS. Then, we introduce \textbf{Dual-Level Curriculum Reinforcement Learning (DCRL)}, a framework that enables the model to autonomously refine its diagnostic logic through structured curricula. Finally, we describe \textbf{Collaborative Reasoning Refinement (CRR)}, an inference-stage method that mitigates single-model biases by synthesizing multi-model perspectives and retrieved evidence.

\subsection{Task Formulation}

\label{sec:task_formulation}

We formulate the diagnostic task as $F_M: C \to (r, \hat{D})$, where model $M$ generates a reasoning trajectory $r$ and ranked candidates $\hat{D} = (\hat{d}_1, \dots, \hat{d}_K)$ from a clinical note $C$, aiming to rank the ground-truth $d^*$ optimally high. 

To mitigate evaluation bias from severe rare disease synonymity, we compiled a multi-source thesaurus consolidating Orphanet\cite{weinreich2008orphanet}, OMIM\cite{amberger2015omim}, and ICD\cite{world1992icd}. This enables our \textbf{synonym-aware matching function}, $\text{match}(\hat{d}, d^*)$, which verifies exact or synonymous matches. Performance is quantified via \textbf{Top-k Recall}.

\subsection{Public Data Resources}
\label{sec:data_resources}
\textbf{Structured Knowledge:} We integrate HPO~\cite{talapova2023human}, Orphanet~\cite{weinreich2008orphanet}, and OMIM~\cite{amberger2015omim} to provide standardized encodings for diseases and phenotypes, along with their associations.

\textbf{Unstructured Knowledge:} Open-access medical literature retrieved from PubMed~\cite{canese2013pubmed} to supplement textual knowledge.

\textbf{Rare Disease Cases:} We utilize the RareArena dataset~\cite{zhao2024rarearena}, comprising 22,896 annotated clinical notes. To ensure balanced coverage, we performed stratified sampling based on disease frequency to partition the data into 1,025 test samples (\textbf{RareArena-Test}), 970 validation samples, and 20,901 training samples (\textbf{RareArena-Train}) used for RareDxCOT generation and RL. Additionally, we collected a subset from MIMIC-IV-Note~\cite{johnson2023mimic} as an out-of-distribution (OOD) benchmark.

\subsection{Knowledge Internalization (RareKnowledgeQA)}
\label{sec:data_knowledge}
A key challenge in rare disease diagnosis for LLMs lies in the fragmented nature of knowledge, split between structured databases and unstructured texts. While structured data ensures factual accuracy and unstructured text provides rich clinical context, our preliminary findings indicated that simply exposing LLMs to raw structured data is insufficient for effective knowledge acquisition.

To address this, we created the \textbf{RareKnowledgeQA} dataset to unify both knowledge types into an LLM-friendly format. We extracted core information—disease definitions, phenotypes, and clinical features—from both sources, and converted it into question-answer pairs using a hybrid approach: rule-based templates for factual precision and LLM-based generation for contextual diversity.

\subsection{Expert-level Reasoning Synthesis (RareDxCOT)}
\label{sec:data_reasoning}
To transition from general reasoning to medical expertise, we constructed \textbf{RareDxCOT}, a dataset of 138,091 verified reasoning trajectories. 

\textbf{Reflection-Enhanced Reasoning Sampling.}
Standard Rejection Sampling (RS) is inefficient as it discards the vast majority of trajectories where the teacher model fails ($\hat{D}$ excludes $d^*$). To address this, we propose RERS for \textbf{Weak-to-Strong} generalization.
Let ${M}_{T}$ be the teacher model (e.g., DeepSeek-R1). For a clinical note $C$, the initial inference process yields a reasoning chain and diagnosis: $(r, \hat{D}) = F_{M_T}(C)$.
We define the reasoning generation process as a conditional mapping to construct a high-quality COT dataset $\mathcal{R}$:

\begin{equation}
\mathcal{R} = 
\begin{cases} 
(r, \hat{D}) & \text{if } \text{match}(\hat{D}, d^*) \\
(r_{ref}, \hat{D}_{ref}) & \text{if } \neg\text{match}(\hat{D}, d^*)
\end{cases}
\end{equation}
\noindent where $(r_{ref}, \hat{D}_{ref}) = F_{M_T}(C, r_{fail}, \mathcal{K}, \mathcal{E})$ denotes the reflection-enhanced trajectory.

The first case represents standard positive sampling where the reasoning successfully leads to the correct diagnosis. The second case denotes our Reflection Mechanism: given a failed reasoning path $r_{fail}$ where the correct diagnosis was missed, we prompt the model to self-correct by synthesizing retrieved knowledge $\mathcal{K}$ and expert feedback $\mathcal{E}$ (insights from other models). This forces the process $F_{M_T}$ to identify logical fallacies in $r_{fail}$ and internalize external knowledge to derive $d^*$, effectively turning "failure" into a dense training signal.

\textbf{Trajectory Refinement.}
To ensure robustness and diversity, we apply two augmentation strategies to the generation of $\mathcal{T}$:
\begin{enumerate}
    \item \textit{Random Information Masking:} We generate $\tilde{C} = \text{Mask}(C)$ by randomly obscuring portions of the input, compelling the model to reason with partial evidence.
    \item \textit{Knowledge-Constrained Filtering:} To eliminate process hallucination, every generated trajectory $r \in \mathcal{T}$ is post-verified against factual triplets extracted from $\mathcal{K}$. Only trajectories satisfying logical consistency and factual accuracy are retained.
\end{enumerate}

\subsection{Dual-Level Curriculum Reinforcement Learning (DCRL)}
\label{sec:training_rl}
Following the injection of knowledge and reasoning via SFT, to move beyond static data dependence and achieve autonomous evolution, we propose the DCRL algorithm. Building upon the DAPO framework~\cite{yu2025dapoopensourcellmreinforcement}, we designed a progressive curriculum from two distinct dimensions, enabling the model to autonomously optimize its reasoning paths.
\paragraph{Dual-Level Curriculum Design}
Unlike mathematical reasoning, where errors primarily stem from insufficient reasoning capabilities, diagnostic failures can also result from a lack of domain knowledge. To decouple and address these challenges, we designed a curriculum operating at two levels:
\begin{itemize}
\item \textbf{Task-level:} Progresses from \textit{Differential Diagnosis} (generating candidates or diagnosing with cues) to \textit{Final Diagnosis} (directly determining a definitive result).
\item \textbf{Case-level:} Samples are stratified into four tiers based on preliminary diagnostic accuracy.
\end{itemize}
This structure enables targeted optimization: combinations of Simple Cases with Hard Tasks focus on mastering multi-step reasoning paradigms, while Hard Cases with Simple Tasks prioritize the recall of disease-specific knowledge.
\paragraph{Scheduling Strategy}
The training curriculum is designed to progressively escalate in both case-level and task-level difficulty. To mitigate catastrophic forgetting, we interleave the current training subset with a small proportion of samples from other difficulty tiers.
\paragraph{Early-Stage Filtering}
Given the severe data scarcity inherent to rare diseases—where many diseases have only one record—we reused RareArena for RL stage. To maximize training efficiency, we filtered out samples that achieved 100\% accuracy during SFT. Our preliminary experiments revealed that reinforcement learning on these data yields negligible performance gains and may even compromise performance.

\begin{table*}[htbp]
\centering
\caption{\textbf{Evaluation on rare disease diagnosis from unstructured clinical notes}. We compare the performance of baseline models, our single models (RareDxR1), and our collaborative refined models (+CRR).}
\begin{tabular}{
@{}l % Model
l % Params
l % Model Domain
l % Reasoning Model
*{3}{c} % RareArena-Test (3 cols)
*{3}{c} % MIMIC-IV-Rare (3 cols)
@{}
}
\toprule
\multirow{2}{*}{\textbf{Model}} & \multirow{2}{*}{\textbf{Params}} & \multirow{2}{*}{\textbf{Domain}} & \multirow{2}{*}{\textbf{RLM}} & \multicolumn{3}{c}{\textbf{RareArena-Test (ID)}} & \multicolumn{3}{c}{\textbf{MIMIC-IV-Rare (OOD)}} \\
\cmidrule(lr){5-7} \cmidrule(lr){8-10}
& & & & \textbf{Top-1} & \textbf{Top-3} & \textbf{Top-10} & \textbf{Top-1} & \textbf{Top-3} & \textbf{Top-10} \\
\midrule
% --- Baseline Models ---
GPT-4o & - & General & No & 41.66 & 50.54 & 56.88 & 29.15 & 33.39 & 37.97 \\
Doubao-1.5-Thinking-Pro & - & General & Yes & 43.71 & 52.88 & 56.78 & 44.58 & 50.34 & 54.75 \\
Gemini-2.5-Flash & - & General & Yes & \underline{44.88} & 53.76 & 57.37 & 38.14 & 45.25 & 48.98 \\
DeepSeek-R1 & 671B & General & Yes & 44.20 & \underline{54.73} & \underline{59.02} & 42.71 & 48.64 & 54.58 \\
DeepSeek-Distill & 14B & General & Yes & 25.76 & 35.12 & 42.15 & 39.83 & 45.76 & 48.31 \\
Qwen3 & 14B & General & Yes & 32.68 & 40.59 & 45.95 & 40.34 & 45.08 & 50.17 \\
Baichuan-M1 & 14B & Medical & Yes & 31.90 & 40.98 & 48.00 & \underline{45.42} & \underline{51.86} & \underline{56.44} \\
HuatuoGPT-o1 & 70B & Medical & Yes & 36.29 & 41.27 & 41.76 & 34.24 & 37.97 & 44.07 \\

\midrule 
% --- Our Models (Single) ---
\multicolumn{10}{l}{\textit{Ours: RareDxR1 (Single Models)}} \\
RareDxR1-DeepSeek & 14B & Medical & Yes & 52.29 & 63.61 & 69.27 & 47.63 & 55.25 & 61.19 \\
RareDxR1-Qwen & 14B & Medical & Yes & \textbf{55.46} & \textbf{66.34} & \textbf{75.32} & \textbf{48.98} & \textbf{57.80} & \textbf{65.25} \\

\midrule 
% --- Our Models (CRR) ---
\multicolumn{10}{l}{\textit{Ours: RareDxR1 + Collaborative Reasoning Refinement (CRR)}} \\
RareDxR1-DeepSeek + CRR & 14B & Medical & Yes & 56.98 & 69.56 & 75.41 & 52.03 & 60.34 & 65.59 \\
RareDxR1-Qwen + CRR & 14B & Medical & Yes & \textcolor{red}{\textbf{60.29}} & \textcolor{red}{\textbf{71.71}} & \textcolor{red}{\textbf{78.73}} & \textcolor{red}{\textbf{53.05}} & \textcolor{red}{\textbf{62.20}} & \textcolor{red}{\textbf{69.83}} \\
\bottomrule
\end{tabular}
\label{result_rarearena_mimic_comparison}
\end{table*}

\subsection{Collaborative Reasoning Refinement (CRR)}
\label{sec:crr}
While RareDxR1 achieves SOTA performance, single models inevitably exhibit biases. To leverage the complementarity between models (Fig.~\ref{fig:collaboration_potential}), we introduce \textbf{CRR}, an inference-time framework mirroring the RERS reflection mechanism.

Let $M$ be the trained RareDxR1 model. Given a clinical note $C$, the process proceeds in steps:

\textbf{1. Initial Diagnosis.} 
It generates an initial reasoning chain $r_{init}$ and a candidate set via $(r_{init}, \hat{D}_{init}) = F_M(C)$.

\textbf{2. Multi-Source Evidence Aggregation.} 
We construct a composite evidence set consisting of external peer opinions and factual verification. Let $\mathcal{M}_{ext}$ be a panel of external models. We gather external diagnoses $\mathcal{E} = \{ F_{M_k}(C) \mid M_k \in \mathcal{M}_{ext} \}$ and retrieve knowledge $\mathcal{K}$ relevant to the union of candidate diseases.

\textbf{3. Refined Inference.} 
Finally, RareDxR1 acts as the lead diagnostician, synthesizing the initial hypothesis with the supplementary evidence to generate the final diagnosis:
\begin{equation}
(r_{final}, \hat{D}_{final}) = F_M(C, r_{init}, \mathcal{E}, \mathcal{K}).
\end{equation}
Crucially, this step utilizes the capability cultivated via RERS—specifically, the ability to critique initial reasoning ($r_{init}$) by weighing conflicting external evidence ($\mathcal{E}$) and factual constraints ($\mathcal{K}$) to correct biases.

\section{Experiments}
\subsection{Experiment Settings}
Our approach was implemented on two general reasoning models, Qwen3-14B\cite{yang2025qwen3} and DeepSeek-Distill-14B\cite{guo2025deepseek}, resulting in RareDxR1-Q and RareDxR1-D. This dual-model setup serves to both validate the general applicability of our method and provide high-quality collaborators for the CRR stage.

Our evaluation of RareDxR1 is threefold. First, we assess its performance on our primary task, Rare Disease Diagnosis from Unstructured Clinical Notes, using the \textbf{RareArena-Test}(1,025 cases, 637 diseases) ~\cite{zhao2024rarearena}  as the in-distribution benchmark and \textbf{MIMIC-IV-Rare} (590 cases, 239 diseases) created from MIMIC-IV-Notes~\cite{johnson2023mimic} for OOD testing. Second, we conduct extensive ablation studies to validate the contribution of each of our proposed modules. Finally, we test generalization on common diseases  on \textbf{DiagnosisArena}\cite{zhu2025diagnosisarena} and diagnosis from structured phenotype inputs on \textbf{RareBench-Public}~\cite{chen2024rarebench}.

Our comparative analysis included mainstream high-performance models spanning open-source and closed-source domains, covering both general-purpose LLMs ~\cite{achiam2023gpt,deepmind2025gemini} and medical reasoning  LLMs ~\cite{chen2024huatuogpt, wang2025baichuan}.

\subsection{Main Results}
Table~\ref{result_rarearena_mimic_comparison} presents the results for rare disease diagnosis from unstructured clinical notes. Following the two-stage training, our RareDxR1-14B model achieves best performance across all  metrics, outperforming numerous models with significantly larger parameter counts. A key indicator of its generalization ability is its superior performance on the MIMIC-IV dataset, despite this data being entirely unseen during training. This robust OOD performance suggests that our approach endows the model with genuine, transferable diagnostic skills applicable across a wide spectrum of rare diseases.

% This highlights the potential of our approach to achieve expert-level complex reasoning without requiring direct human annotation, demonstrating a scalable path toward building highly capable diagnostic models.

As shown in Table~\ref{tab:freq_compact}, to validate whether RareDxR1 acquires generalizable reasoning rather than rote memorization, we stratified the test cases by the frequency of disease in the train set. RareDxR1-Q consistently outperforms the DeepSeek-R1-671B baseline. Crucially, in the \textbf{Zero-shot} setting, our model achieves a \textbf{50.00\%} Top-10 Recall (vs. 28.85\% for baseline). This confirms that RareDxR1 has internalized the meta-logic of differential diagnosis, enabling it to infer correct diagnoses even without prior exposure to specific disease samples.

\begin{table}[t]
    \centering
    \small % 极小字号
    \setlength{\tabcolsep}{1.5pt} % 极小列间距
    \renewcommand{\arraystretch}{0.9} % 压缩行高
    \caption{\textbf{Top-1/10 Recall} by disease frequencies in train set. "Freq=0" denotes unseen diseases.}
    \label{tab:freq_compact}
    \begin{tabular}{@{}lccccc@{}}
    \toprule
    \multirow{2}{*}{\textbf{Group (Train-Freq)}} & \multirow{2}{*}{\textbf{\#Case}} & \multicolumn{2}{c}{\textbf{DS-R1-671B}} & \multicolumn{2}{c}{\textbf{Ours-Q (14B)}} \\ 
    \cmidrule(lr){3-4} \cmidrule(l){5-6}
     &  & \textbf{Top-1} & \textbf{Top-10} & \textbf{Top-1} & \textbf{Top-10} \\ \midrule
    Zero-shot (0) & 52 & 19.23 & 28.85 & \textbf{28.85} & \textbf{50.00} \\
    Few-shot ($<$19) & 544 & 40.99 & 56.62 & \textbf{43.38} & \textbf{67.46} \\
    Many-shot ($\ge$19) & 429 & 51.28 & 65.73 & \textbf{69.23} & \textbf{88.34} \\ \bottomrule
    \end{tabular}
\end{table}

\begin{figure}[htbp]
    \centering
    \includegraphics[width=\columnwidth]{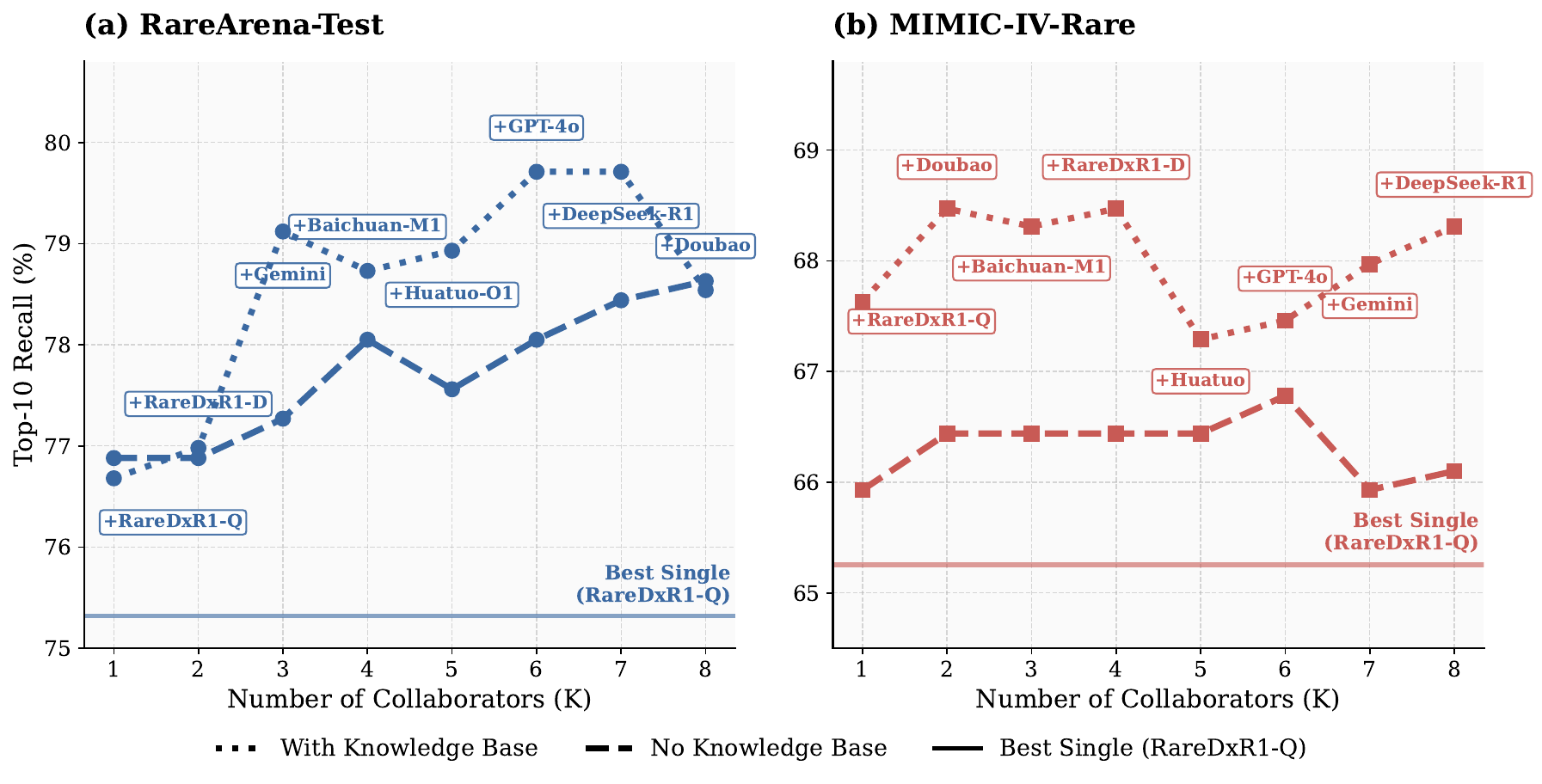}
\caption{
    \textbf{Analysis of  CRR Performance.} 
    Performance of our method as collaborators are added with or without retrieved knowledge. $K=1$ represents the model's self-refinement using retrieved knowledge.
}
    \label{fig:collaboration_potential}
\end{figure}

To investigate the theoretical basis for CRR's effectiveness, we analyzed the collaborative potential of multiple models as shown in Fig.~\ref{fig:collaboration_potential}. This confirms the existence of complementary diagnostic strengths among models and validates that synthesizing diverse perspectives, as CRR does, is a promising direction for enhancing diagnostic accuracy.

% 插入图表

\subsection{Ablation Studies}
To validate the contribution of each proposed component, we conducted an ablation study, with results presented in Table~\ref{tab:ablation_study}. The findings reveal a clear, step-wise performance gain, confirming the efficacy of our multi-stage training strategy.
The most significant performance leap originates from our SFT stage, which demonstrates the critical importance of domain-specific adaptation. Within this stage, the inclusion of data from our Reflection-Enhanced Reasoning Sampling pipeline provides a distinct advantage, validating our hypothesis that learning from corrected errors enhances reasoning quality. Subsequently, applying Reinforcement Learning with our dual-level curriculum further hones the model's capabilities, delivering the final and most substantial boost in performance. 

Notably, a "weak-to-strong" generalization phenomenon was observed. Even though powerful models like DeepSeek-R1-671B served as teachers in the generation of our RareDxCOT, our model, after only SFT, substantially surpassed its teachers' performance. We attribute this significant improvement to our data generation strategies, such as knowledge-constrained filtering and reflection-enhanced sampling, which effectively distill and refine the teacher's capabilities.

\begin{table}[t]
\small
\centering
\setlength{\tabcolsep}{1.5pt}
\caption{Ablation study on the contributions of SFT and RL on the RareArena-Test.}
\begin{tabular}{lccc}
\toprule
\textbf{Model Configuration} & \textbf{Top-1} & \textbf{Top-3} & \textbf{Top-10} \\
\midrule
(A) Base Model (DeepSeek-Distill) & 25.76 & 35.12 & 42.15 \\
\midrule
\multicolumn{4}{l}{\textit{+SFT Only}} \\
(B) SFT (w/o RERS Data) & 45.85 & 56.20 & 60.78 \\
(C) SFT (w/ RERS Data) & 47.12 & 56.98 & 62.15 \\
\midrule
\multicolumn{4}{l}{\textit{+RL}} \\
(D) RL Only & 30.24 & 43.02 & 52.10 \\
(E) SFT + RL (w/o Curriculum) & 49.56 & 60.49 & 65.56 \\
(F) \textbf{SFT + RL (w/ Curriculum)} & \textbf{52.29} & \textbf{63.61} & \textbf{69.27} \\
\bottomrule
\end{tabular}
\label{tab:ablation_study}
\end{table}

%__________

\begin{table}[tb]
\centering
\small
\setlength{\tabcolsep}{1mm}
\caption{Zero-shot accuracy of RareDxR1 on auxiliary tasks, demonstrating strong generalization across different data modalities (notes vs. phenotypes) and disease domains}
\begin{tabular}{lcc}
\toprule
\multirow{2}{*}{\textbf{Model}} & \textbf{RareBench-Public} & \textbf{DiagnosisArena} \\
\cmidrule(lr){2-2} \cmidrule(lr){3-3} 
& \textit{Phenotypes, Rare} & \textit{Notes, Common} \\ \midrule % Added \\ here and fixed content
\multicolumn{3}{l}{\textit{Specialized Method (Closed-set)}} \\
PhenoBrain & 30.40 & -- \\
\midrule
\multicolumn{3}{l}{\textit{Large-scale LLMs (Open-domain)}} \\
DeepSeek-R1-671B & \textbf{37.92} & \textbf{45.57} \\
Gemini-2.5-Flash & 35.51 & 37.48 \\
Doubao-1.5-Thinking-Pro & 31.96 & 30.93 \\
GPT-4o & 30.01 & 24.04 \\
\midrule
\multicolumn{3}{l}{\textit{Comparable-sized LLMs (Open-domain)}} \\
Baichuan-M1-14B & 30.01 & 22.30 \\
HuatuoGPT-o1-70B & 29.44 & 18.69 \\
DeepSeek-Distill-14B & 30.93 & 27.76 \\
\textbf{RareDxR1-D-14B (Ours)} & \textbf{37.00} & \textbf{35.73} \\
\bottomrule
\end{tabular}
\label{tab:generalization_tasks}
\end{table}

\subsection{Cross-Domain Generalization Ability}
To validate that our training cultivates a truly generalizable medical reasoning ability, we evaluated RareDxR1's zero-shot performance on two distinct auxiliary tasks: Rare Disease Diagnosis from Structured Phenotypes and Common Disease Diagnosis from Unstructured Notes. The consolidated results are presented in Table~\ref{tab:generalization_tasks}.

On RareBench-Public, RareDxR1 demonstrates robust cross-modal transfer without phenotype training. Despite operating in a more challenging open-domain setting compared to PhenoBrain's closed-set paradigm, it outperforms this specialized SOTA and rivals top-tier LLMs.

Furthermore, results on the DiagnosisArena benchmark demonstrate that RareDxR1 generalizes effectively to common diseases without suffering from catastrophic forgetting. Notably, it surpasses its base model (DeepSeek-Distill), indicating that our training strategy enhances diagnostic capabilities.

\section{Conclusion}
RareDxR1 enables reasoning-centric rare disease diagnosis from unstructured notes. Through knowledge-reasoning dataset construction and dual-level curriculum reinforcement learning, our model outperforms larger baselines. Furthermore, our Collaborative Reasoning Refinement facilitates effective joint reasoning with external experts, underscoring RareDxR1's potential as a powerful tool for clinicians. Meanwhile, rigorous clinical validation and human oversight remain essential to ensure patient safety in real-world applications.

\bibliographystyle{IEEEtran}
\bibliography{icme2026references}
\end{document}